\documentclass[runningheads,a4paper]{llncs}

\usepackage[T1]{tipa}
\usepackage[english]{babel}
\usepackage{courier}
\usepackage{amstext}
\usepackage{times}
\usepackage{helvet}
\usepackage{courier}
\usepackage[cmex10]{amsmath}
\usepackage{amssymb}
\usepackage{latexsym}
\usepackage{eurosym}
\usepackage{wasysym}
\usepackage{array}

\makeatletter
\newif\if@restonecol
\makeatother

\usepackage{etex}
\usepackage{tikz}
\usepackage[T1]{fontenc}
\usepackage{pslatex}
\usepackage[arrow, matrix, curve]{xy}
\usepackage{epsfig}
\usepackage{pstricks,egameps}
\usepackage{hyperref, url}
\usepackage[algochapter, boxruled, linesnumbered]{algorithm2e}

\begin{document}

\title{Business Negotiation Definition Language}
\titlerunning{Business Negotiation Definition Language}

\author{Rustam Tagiew\inst{1}}
\authorrunning{Rustam Tagiew}

\institute{Alumni of TU Bergakademie Freiberg and Uni Bielefeld\\
\email{rustam@tagiew.de}, \texttt{www.tagiew.de}}
\maketitle

\begin{abstract} 
The target of this paper is to present an industry-ready prototype software for general game playing. This software can also be used as the central element for experimental economics research, interfacing of game-theoretic libraries, AI-driven software testing, algorithmic trade, human behavior mining and simulation of (strategic) interactions. The software is based on a domain-specific language for electronic business to business negotiations -- SIDL3.0. The paper also contains many examples to prove the power of this language.  
\end{abstract}
\section{Motivation}
\indent This paper describes the scientific background of a ready implemented prototype software available under github.com/Yepkio/sidl \cite{yepkio2019}.\\
\indent The issue discussed in this paper provides improvements to many fields simultaneously and facilitates their interconnections. Since the extent of this issue is so over-sized to almost any reader non-proficient or only partially proficient in the overall carpet of research fields, the paper proceeds along the leitmotif of a business application. The concrete business applications are the electronic business to business (B2B) negotiations as Friedrich Michael described in his paper \cite{friedrich}.\\
\indent These electronic B2B negotiations (EB2BN) have equilibria in game-theoretic terms, which should be observed in case of rationality and perfect reasoning. Although this analytical solution by game-theoretic math is rarely the case in real world, certain convergence towards this solution exists. Game-theoretic solution also provides a rough estimate of the soundness for a certain EB2BN. For instance, when a buying company sets up a complicated EB2BN with multiple sellers, it traces certain goals of reducing the price or just of receiving and spreading information. The software is desired to be able to make an automatic game-theoretic analysis of an EB2BN defined in a domain-specific language. The seminal work solving this kind of problems is Gala \cite{gala}.\\
\indent EB2BNs are implemented by computer systems. General Game Playing (GGP) means development of AI systems, which are based on a domain-specific language for games and therefore can deals with games in general terms \cite{ggp}. For EB2BNs, this means development of a business negotiation engine, which runs on a domain-specific language for EB2BNs. In GGP, it is called game management and calculates the game states for traversal.\\
\indent A very interesting topic for EB2BN is algorithmic trade. The selling companies hereby employ automated negotiators. The automated negotiators aka agents receive the definition of EB2BN like in GGP in advance. Then, the  agents can make intelligent decisions in favor of their shepherd.\\
\indent Agents will not be the case in the beginning. Human employees of the sellers will be put in front of a user interface to the EB2BN. Those humans have certain typically human behavior patterns, which are studies using the data from the laboratory experiments. And those experiments are carried out by experimental economists. A domain-specific language would greatly facilitate transfers from field to laboratory and back. Human behavior mining would greatly profit from clearly defined EB2BN setups.\\
\indent Current industry solutions do not use any GGP style engines for EB2BNs. Current EB2BN systems are hard-coded. Programming of systematic testing of these software pieces requires deep knowledge of the concrete EB2BNs. Testing software, which is based on a domain-specific language would be much easier to handle.\\
\indent And finally, offline simulation of EB2BNs is hardly imaginable without a domain-specific language. The goal of an offline simulation is to test a concrete EB2BN with humans or/and agents. If this EB2BN does return the needed results, the simulation should be easily rerun with an adjusted EB2BN.\\
\indent Section \ref{related} briefly summarizes prehistory of this development. Section \ref{langdef} is hopefully the easiest way to present the formal background of SIDL3.0 -- a language usable for EB2BN. Section \ref{chronon} presents the state transition calculating algorithm for SIDL3.0. Examples of game definitions in SIDL3.0 are given in section \ref{samples}.\\
\section{Related Work}
\label{related}
\indent A good and still prevailing review \cite{tagiew2011beyond} presents languages for game-theoretic problems. Languages and frameworks for setting up experiments in behavioral economics also are numerous \cite{Balietti2017}.\\
\indent Strategic Interaction Definition Language (SIDL) has a decade long prehistory. {\it Strategic interaction} is a more clear term for a game as meant in Game Theory. SIDL was first introduced in 2009 \cite{tagiewggma} as a universal language for games of imperfect information. Then, it was significantly improved as SIDL2.0 \cite[p.98--105]{tagiewPhD}. This paper presents SIDL3.0, which is minor improvement of SIDL2.0 adding unlimited spaces for actions.\\
\section{Language Definition}
\label{langdef}
\indent SIDL \textipa{['zaId@l]} is a first-order logic definition of a game. SIDL structure is derived from {\it STRIPS} formalism \cite{strips} and {\it situation calculus} \cite{situationcalculus}. The formal definition of SIDL3.0 is given in Def.\ref{e.sidl2}. Like in the language for planning tasks {PDDL} \cite{pddl}, SIDL describes the state of the world by a set of {\it facts}. From SIDL2.0 on, these facts are lists of literals, called {\it words}. {\it Closed world assumption} applies for SIDL. In chess for instance, following facts mean that white knight is on c4, black queen is on e6 and nothing else is on the board:
\begin{verbatim}
 [white, knight, c, 4]
 [black, queen, e, 6]
\end{verbatim}
According to chess rules, there are multiple squares on this board, where the knight and the queen can move in next turn from this {\it initial state} (Def.\ref{e.sidl2}, 2). Every of such moves is an {\it action} in SIDL (Def.\ref{e.sidl2}, 3). Together with a definition of a state transition (Def.\ref{e.sidl2}, 4) to every of these actions or their combinations, it is already a definition of a planning task, if 'players' and 'switches' are blanked out.   
\begin{definition}[SIDL3.0]
\label{e.sidl2}
A game description in SIDL3.0 consists of following elements, whereby $\Sigma$ is a finite set. $\Sigma$ consists of a finite set of symbols and a finite subset of real numbers:
\begin{enumerate}
 \item $S\subset\{w\circ m\colon w\subset\Sigma^*\wedge m\in\mathbb{R}^{\vert N\vert}\}$ a set of states. Every state is set of words consisting of symbols $\Sigma$ and an assignments for players' accounts.
 \item $s_{go}\in S$ initial state. 
 \item $A\subset\Sigma^*$ a set of actions, every action is a word consisting of symbols from $\Sigma$.  
 \item $\rhd\colon S\times A^{m}\rightarrow S$ Definition of state transitions for a SIDL-automaton. $m$ is the number of simultaneously applicable actions, which is bigger than $0$ and does not exceed the number of legal switches.
 \item $N\subset\Sigma^*$ a set of players, which are words consisting of symbols from $\Sigma$.
 \item $I\subset\Sigma^*$ a set of switches.
 \item $L\colon I\times\Sigma^{**}$ a truth function to define the legality of a switch. Terminal states have an empty set of legal switches.
 \item $P\colon I\times\Sigma^{**}\rightarrow A^*$ a function to define a set of actions for every switch depending on state. This function is bijective.
 \item $O\colon\{i\circ w\mapsto n\colon w\subset \Sigma^*\wedge i\in I\wedge n\in(N\cup(\mathbb{R}^1_0)^{\vert P(i,w)\vert})\}$ an attribution of a player or of a distribution to a switch depending on state.
 \item $H\colon \Sigma^*\times N^*$ a truth function for hidden words. 
 \item Following applies:\begin{multline}
    \nonumber
    \forall n\in N\colon\forall s, s'\in S\colon\forall l\in I\colon\\
    (\{w\in s\colon\neg H(w,n)\}=\{w\in s'\colon\neg H(w,n)\})\wedge (n = O(l,s))\Rightarrow\\
    n = O(l,s')\wedge(P(l,s)=P(l,s'))\wedge(L(l,s)\Leftrightarrow L(l,s'))
\end{multline}
\end{enumerate}
\end{definition}
\indent The elements 5-11 of Def.\ref{e.sidl2} are needed, since we have more than one 'planner' in a game. Element 5 defines players as words similar to facts -- it simplifies the definition of general rules.\\
\indent Elements 6-8 define the sets of mutually exclusive actions as {\it switches}. A switch is basically a set of actions, from where only one can be applied in a state transition. If game rules require multiple simultaneous actions, then actions from multiple switches should be used. There is also a function, which defines {\it legal}ity of a switch depending on the game state. The set of actions belonging to a switch also depends on the game state -- it adds casual flexibility to SIDL3.0.\\ 
\indent Element $O$ defines either an ownership of a switch by player or an assignment to a probability distribution over the available actions. This probability distribution is needed for games, which include random events. Like the set of actions itself, the distribution also depends on the game state.\\
\indent The game-theoretic notion of {\it imperfect information} should not be confused with {\it incomplete information}. Not seeing parts of the game state is called imperfect, and not seeing parts of the game rules is called incomplete. $H$ defines imperfect information for SIDL by hiding facts from players. This is a delicate matter in SIDL -- the hidden facts should neither be revealed by legality of switches, nor by the sets of actions from switches for a certain player (Def.\ref{e.sidl2}, 11). Otherwise a player might deduce the hidden facts -- they are not hidden anymore.\\
\indent Finally, to predict rational or rationally motivated behavior of players given the game rules, we need some kind of preferences being defined in the game rules. This is done by 'special facts' -- {\it the payers' accounts}. Every change of the game state might change them by {\it payoffs}. A rational player is expected to maximize his/her account. It can also be a case that the real playoffs are not known.\\   
\indent The syntax of SIDL is based on ISO Prolog \cite{isoprolog}. ISO Prolog is a logic programming language -- it has in its pure form neighter loops nor conditionals. There are rules consisting of a rule head and of a rule body. Rule body is a condition for the rule head. Following language elements are available for game description in SIDL3.0:
\begin{itemize}
 \item All ISO Prolog operators for lists' and numbers' manipulations are allowed. Operators \textit{game}, \textit{init}, \textit{hidden}, \textit{legal}, \textit{switch}, \textit{unlimited}, \textit{owned}, \textit{default}, \textit{do} and \textit{payoff} are keywords of SIDL3.0 and only allowed in rule heads. \textit{player}, \textit{fact}, \textit{create}, \textit{delete}, \textit{tocreate}, \textit{todelete} and \textit{does} also are keywords and are only allowed in rule bodies.
 \item \textit{game($NAME$)} -- $NAME$ is the name of the game.
 \item \textit{init($W$):-$Condition$} -- such rules define the words of the initial state. Every word \textit{$W$ = [a, $\ldots$]} is represented as a list.
 \item \textit{player($N$)} -- this operator returns a valid player $N$.
 \item \textit{fact($W$)} -- this returns valid word $W$.
 \item \textit{hidden($W$, $P$):-$Condition$} -- word $W$ is hidden for player $P$.\\
$Condition$ may contain atoms with operator \textit{player}. 
 \item \textit{legal($I$):--$Condition$} -- list $I$ is legal switch in the actual state. $Condition$ may contain atoms with operator \textit{player} and \textit{fact}.
 \item \textit{owned($I$, $D$):-$Condition$} -- switch $I$ is attributed to a player or to a distribution $D$. A distribution is either a list of probabilities or an expression \textit{equal($X$)}, which is a uniform distribution over $X$ actions. $Condition$ may contain atoms with operators \textit{player} and \textit{fact}.
 \item \textit{switch($I$, $A$):-$Condition$} -- list $A$ is one of possible actions for switch $I$. $Condition$ may contain atoms with operators \textit{player} and \textit{fact}.
 \item \textit{unlimited($I$, $T$):-$Condition$} -- if a list of actions for a switch $I$ is unreasonable to be created due to its size, $T$ can be defined as its template structure by this operator. $Condition$ may contain atoms with operators \textit{player} and \textit{fact}.
 \item \textit{create($W$)} -- word $W$ will be added in next turn.
 \item \textit{delete($W$)} -- word $W$ will be deleted in next turn.
 \item \textit{tocreate($W$)} -- returns a word $W$, which will be added in next turn.
 \item \textit{todelete($W$)} -- returns a word $W$, which will be deleted in text turn.
 \item \textit{does($I$, $A$)} -- returns the action $A$ for the switch $I$.
 \item \textit{do($A$):-$Condition$} -- definition of the action $A$. $Condition$ may contain atoms with operators \textit{player}, \textit{fact},  \textit{create}, \textit{delete} and \textit{does}. 
 \item \textit{default($I$, $A$):-$Condition$} -- default action $A$ of the switch $I$. $Condition$ may contain operators \textit{player} and \textit{fact}.
 \item \textit{init($N$, $M$):-$Condition$} -- player accounts in the initial state. List $N$ is a player and $M$ is its account balance.
 \item \textit{payoff($N$, $R$):-$Condition$} -- payoff $R$ will be added to the account of player $N$. $Condition$ may contain atoms with operators \textit{player}, \textit{fact}, \textit{tocreate}, \textit{todelete} and \textit{does}.
\end{itemize}
\section{Chronons versus Interrupts}
\label{chronon}
\indent Based on a preliminary decision of allowing or denying simultaneous actions by multiple players, there are two ways of game management based on SIDL. The first is based on chronons as time quanta between state transitions, during which all players' actions are considered simultaneous. The second is an interrupt based approach with every valid action triggering a state transition and no simultaneous actions. For the interrupt based approach, $m$ from element 4 of Def.\ref{e.sidl2} equals always $1$. The chronon based approach offers obviously more advantages -- the interrupt based approach is skipped from consideration in this paper.\\
\indent Alg.\ref{routine11} shows the game management algorithm for the chronon based approach. \textbf{lcall} stands for calls into the Prolog engine. {\it assert} and {\it retract} add and delete facts from the game state. The algorithm runs in a loop as long as there are legal switches. Every loop run lasts one chronon.\\ 
\begin{footnotesize}
\begin{algorithm}[pt]
\caption{Game management based on SIDL3.0 \cite[p.106]{tagiewPhD}.}
\label{routine11}
\ForEach{F in \textbf{lcall}(init(F)).solutions}{
\textbf{lcall}(assert(fact(F)))\;
}
\ForEach{(N, M) in \textbf{lcall}(init(N, M)).solutions}{
\textbf{lcall}(assert(account(N, M)))\;
}
\While{\textbf{lcall}(legal(\_))}{
\While{chronon\_not\_expired}{
  assert\_does\_for\_a\_valid\_incomming\_command\;
}
\ForEach{(I,D) in \textbf{lcall}(legal(I)$\wedge$owned(I,D)$\wedge$dist(D)).solutions}{
  assert\_randomly\_does(I,D)\;
}
\ForEach{(I,A) in \textbf{lcall}(legal(I)$\wedge$(does(I,A)$\vee$default(I,A))).solutions}{
  \textbf{lcall}(do(A))\;
}
\ForEach{(N,P) in \textbf{lcall}(goal(N,P)).solutions}{
  \textbf{lcall}(retract(account(N,M))$\wedge$assert(account(N,M+P)))\;
}
\ForEach{(F) in \textbf{lcall}(tocreate(F)$\vee$todelete(F)).solutions}{
  send\_unhiden\_changes\_to\_players(F)\;
}
\ForEach{(N,M) in \textbf{lcall}(account(N,M)).solutions}{
  send\_to\_players(N,M)\;
}
\ForEach{(F) in \textbf{lcall}(todelete(F)).solutions}{
  \textbf{lcall}(retract(todelete(F))$\wedge$retract(fact(F)))\;
}
\ForEach{(F) in \textbf{lcall}(tocreate(F)).solutions}{
  \textbf{lcall}(retract(tocreate(F))$\wedge$assert(fact(F)))\;
}
\textbf{lcall}(retractall(does(\_,\_)))\;
}
end\_the\_game\;
\end{algorithm}
\end{footnotesize}
\indent There are cases in business application, where the rules have to be changed during a game run \cite{friedrich}. For instance, new players have to be added. This extraordinary situation equals to a restart of a complete game, reassignment of agents into the new player roles and sending them the new game rules.\\
\section{Examples}
\label{samples}
\indent Following examples will present variants of definitions for commonly known games. Some of less important details of the game rules will not be commented. The reader is welcome to dive into the code and maybe find better definitions for those games.\\   
\indent In game Nim, players have to subtract in each turn between $1$ and $3$ items from a pool. Players rotate and the one, who is last at subtracting loses.
\begin{verbatim}
name(nim).
init([alice, 10]).
init([alice], 0.0).
init([bob], 0.0).
legal([main]) :- fact([_, I]), I > 0.    
switch([main], [T]) :-
    fact([_, I]), M is min(I, 3), between(1, M, T). 
switch([main], [wait]).
owned([main], [A]):- fact([A, _]). 
default([main], [1]).
do([wait]).                                               
do([T]):- 
    fact([A, I]), NI is I - T, player([B]), 
    not(A = B), delete([A, I]), create([B, NI]).
payoff([A], 1.0):- tocreate([A, 0]).
payoff([A], -1.0):- tocreate([B, 0]), not(A = B).
\end{verbatim}
\indent Muddy Children Puzzle (MCP) is a frequently cited logic puzzle. In this definition, there are $5$ children and $n$ of then get muddy faces by chance. Every child sees other children's faces, but he/she does not see the own one. Later, children stand in a row -- they are given an infinite sequence of opportunities to step forward in order to signal own muddy face. If own face is muddy, every missed opportunity for stepping forward is bad ($-1$). If own face is not muddy, stepping forward is much worse ($-10000$). All $n$ muddy children step forward at $n$-th opportunity in case of rational and perfectly reasoning children.     
\begin{verbatim}
name(mcp).
children([alice, bob, charly, david, eric]).
makedirty([]).
makedirty([C | Cs]):- create([dirty, C]), makedirty(Cs).
init([start]).
init([C], 0.0):- children(Cs), member(C, Cs).
hidden([dirty, C], [C]):- player([C]).
legal([dirt]):- fact([start]).
legal([C]):-
    player([C]), not(fact([start])), not(fact([_, stepped])).
switch([dirt], [dirt | SCs]):-
    children(Cs), getsubset(SCs, Cs), not(SCs = []).
switch([C], [C, S]):-
    player([C]), member(S, [stay, step]).
owned([dirt], equal(N)):-
    children(Cs), length(Cs, L), N is (2^L)-1.
owned([C], [C]):- player([C]).
default([C], [stay]):- player([C]).
do([ dirt | DCs ]):-
    fact([start]), delete([start]), makedirty(DCs).
do([_, stay]).
do([C, step]):- not(fact([start])), create([C, stepped]).
payoff([C], -1):-
    not(fact([start])), not(tocreate([C, stepped])).	
payoff([C], 100.0):-
    tocreate([C, stepped]), fact([dirty , C]).
payoff([C], -10000.0):-
    tocreate([C, stepped]), not(fact([dirty , C])).
\end{verbatim}
\indent This is the well-known Rock Paper Scissors in SIDL3.0. There are $10$ rounds in this game definition. Every round lasts $3$ chronons. Already made actions from the previous round are shown to everybody. The default action for both players is Rock.
\begin{verbatim}
name(rps).
gestures([rock, paper, scissors]).
beats(paper, rock).
beats(rock, scissors).
beats(scissors, paper).
opponent(role1, role2).
opponent(role2, role1).
giventime(3).
givenrounds(10).
gesture(P, G):-
    not(does([P], _)), fact([chosen, P, G]).
gesture(P, G):-
    does([P], [P, wait]), fact([chosen, P, G]).
gesture(P, G):- does([P], [P, G]).
regesture(P, G):-
    gesture(P, G).
regesture(P, G):-
    gesture(P, O), not(G = O),
    delete([chosen, P, O]), create([chosen, P, G]).
madegestures:-
    gesture(role1, G1), create([made, role1, G1]),
    gesture(role2, G2), create([made, role2, G2]).
init([gameon]).
init([chosen, P, rock]):- opponent(P, _).
init([rounds, 10]):- givenrounds(10).
init([timer, X]):- giventime(X).
init([P], 0.0):- opponent(P, _).
hidden([chosen, P, _], [O]):- opponent(P, O).
legal([P]):- fact([gameon]), player([P]).
legal([timer]):- fact([timer, X]), X > 0.
legal([round]):- fact([timer, 0]), fact([gameon]).
switch([P], [P, wait]):- player([P]).
switch([P], [P, G]):-
    player([P]), gestures(Gs), member(G, Gs).
switch([timer], [timer]).
switch([round], [round]).
owned([P], [P]):- player([P]).
owned([timer], equal(1)).
owned([round], equal(1)).
do([_, wait]).
do([P, G]):-
    player([P]), fact([chosen, P, O]),
    not(O = G), 
    delete([chosen, P, O]), create([chosen, P, G]).
do([timer]):-
    fact([timer, X]), X > 0, XMM is X - 1,
    delete([timer, X]), create([timer, XMM]).
do([round]):-
    fact([rounds, X]),
    X > 1,
    delete([timer, 0]), delete([rounds, X]),
    NX is X - 1,
    create([rounds, NX]),
    giventime(T), create([timer, T]),
    madegestures.	    
do([round]):-
    fact([rounds, 1]), madegestures, delete([gameon]).
payoff([P], 1.0):-
    does([round], [round]), gesture(P, G1),
    opponent(P, O), gesture(O, G2), beats(G1, G2).
payoff([P], 0.0):- opponent(P, _).
\end{verbatim}
\indent This example EB2BN has a starting price and players have to overbid it. This definition adds all bids to game state and calculates the highest of them for every new state. Operator \textit{unlimited} is used to define bids as real numbers.    
\begin{verbatim}
bidders([alice,bob,clara,david]).
name(priceNegotiation).
init([startprice, 10.0]).
init([A], 0.0):- bidders(Rs), member(A,Rs).
legal(R):- player(R).    
unlimited(R, [wait]):- player(R).
unlimited([A], [A, (price,double)]):- player([A]).      
leadingprice(B):-
    not(fact([bid,_,_])), fact([startprice,B]).
leadingprice(B):-
    fact([bid,_,B]), findall(P, fact([bid,_,P]),Ps),
    maxmember(Ps,B).    
owned(R, R):- player(R).     
switch([A], [A, T]):- player([A]), leadingprice(B), T > B.   
switch(R, [wait]):- player(R).
default(R, [wait]):- player(R).
do([wait]).               
do([A, P]):- create([bid, A, P]).
payoff(R, 0.0):- player(R).
\end{verbatim}
\indent Chess has four types of actions -- move, take, castle and promote. Castling can be left or right. The precondition for castling can be irreversibly invalidated for both players. Therefore in addition to the positions of men, chess needs facts representing the castling precondition. This definition of chess does not contain a definition of a draw nor a limit on turns.
\begin{verbatim}
name(chess).
opposite(white, black).
opposite(black, white).
xaxis([a,b,c,d,e,f,g,h]).
direction(white, 1).
direction(black, -1).
forpromotion([queen, knight, rook, bishop]).
directions(rook, [[1, 0], [0, 1], [-1, 0], [0, -1]]).
directions(bishop, [[1, 1], [1, -1], [-1, 1], [-1, -1]]).
directions(queen, A):- 
    directions(rook, R), directions(bishop, B), append(R, B, A).
position(knight, X, Y):- member(X, [-2, 2]), member(Y, [-1, 1]).
position(knight, X, Y):- member(Y, [-2, 2]), member(X, [-1, 1]). 
position(king, X, Y):- directions(queen, A), member([X, Y], A).
xvalue(Pre, Next, Inc):- 
    xaxis(Xs), nth1(I, Xs, Pre), II is I + Inc, nth1(II, Xs, Next).
yvalue(Y):- integer(Y), Y < 9, Y > 0.
yvalue(Pre, Next, Inc):- Next is Pre + Inc, yvalue(Next).    
row(white, 1, 1).
row(white, 2, 2).
row(black, 8, 1).
row(black, 7, 2).
man(rook, [a, h], 1).
man(knight, [b, g], 1).
man(bishop, [c, f], 1).
man(queen, [d], 1).
man(king, [e], 1).
man(pawn, Xs, 2):- xaxis(Xs).
promote(Color, Y, Man):- 
    opposite(Color, Opp), row(Opp, Y, 1), 
    forpromotion(HMs), member(Man, HMs).
init([Color, Man, X, Y]):- 
    row(Color, Y, Row), man(Man, Xs, Row), member(X, Xs).
init([white]).  
init([white, fortifiable]).
init([black, fortifiable]).
init([white], 0.0).
init([black], 0.0).
legal([C]):- fact([C, king, _, _]), player([C]), fact([C]).
contrline(_, X, _, [DX, _], []) :- not(xvalue(X, _, DX)).
contrline(_, _, Y, [_, DY], []) :- not(yvalue(Y, _, DY)).
contrline(Color, X, Y, [DX, DY], []) :-  
    xvalue(X, X1, DX), yvalue(Y, Y1, DY), fact([Color, _, X1, Y1]).    
contrline(Color, X, Y, [DX, DY], [[X1, Y1]]):-
    xvalue(X, X1, DX), yvalue(Y, Y1, DY),
    opposite(Color, Opp), fact([Opp, _, X1, Y1]).    
contrline(Color, X, Y, [DX, DY], [[X1, Y1] | Line]):-
    xvalue(X, X1, DX), yvalue(Y, Y1, DY),
    not(fact([_, _, X1, Y1])),
    contrline(Color, X1, Y1, [DX, DY], Line).  
controlled(Color, pawn, X, Y, X1, Y1):-
    fact([Color, pawn, X, Y]),
    direction(Color, I),
    yvalue(Y, Y1, I), member(D, [1, -1]), xvalue(X, X1, D).
controlled(Color, Man, X, Y, X1, Y1):-
    directions(Man, Ds),
    fact([Color, Man, X, Y]),
    member(D, Ds),
    contrline(Color, X, Y, D, NP),
    member([X1, Y1], NP).
controlled(Color, Man, X, Y, X1, Y1):-
    position(Man, DX, DY),
    fact([Color, Man, X, Y]),
    xvalue(X, X1, DX), yvalue(Y, Y1, DY), 
    not(fact([Color, _, X1, Y1])).   
switch([Color], [Color, pawn, X, Y, X, Y2]):-
    fact([Color, pawn, X, Y]), row(Color, Y, 2),
    direction(Color, I),
    yvalue(Y, Y2, I * 2), yvalue(Y, Y1, I),
    not(fact([_, _, X, Y1])),
    not(fact([_, _, X, Y2])).
switch([Color], [Color, pawn, X, Y, X, Y1]):-
    fact([Color, pawn, X, Y]),
    direction(Color, I), yvalue(Y, Y1, I),
    not(promote(Color, Y1, _)),
    not(fact([_, _, X, Y1])).
switch([Color], [Color, pawn, X, Y, X, Y1, Man]):-
    fact([Color, pawn, X, Y]),
    direction(Color, I), yvalue(Y, Y1, I),
    promote(Color, Y1, Man),
    not(fact([_, _, X, Y1])).
switch([Color], [Color, pawn, X, Y, X1, Y1, Man]):-
    controlled(Color, pawn, X, Y, X1, Y1),
    opposite(Color, Opp), fact([Opp, _, X1, Y1]),
    promote(Color, Y1, Man).
switch([Color], [Color, pawn, X, Y, X1, Y1]):-
    controlled(Color, pawn, X, Y, X1, Y1),
    opposite(Color, Opp), fact([Opp, _, X1, Y1]),
    not(promote(Color, Y1, _)).  
switch([Color], [Color, Man, X, Y, X1, Y1]):-
    controlled(Color, Man, X, Y, X1, Y1),
    not(Man = pawn), not(Man = king).
switch([Color], [Color, king, X, Y, X1, Y1]):-
    controlled(Color, king, X, Y, X1, Y1),
    opposite(Color, Opp),
    not(controlled(Opp, _, _, _, X1, Y1)).
switch([Color], [Color, castle, right, Y]):-
    fact([Color, fortifiable]),
    fact([Color, rook, h, 1]), 
    row(Color, Y, 1), not(fact([_, _, f, Y])),
    not(fact([_, _, g, Y])), opposite(Color, Opp),
    not(controlled(Opp, _, _, _, f, Y)),
    not(controlled(Opp, _, _, _, g, Y)).
switch([Color], [Color, castle, left, Y]):-
    fact([Color, fortifiable]),
    fact([Color, rook, a, 1]), row(Color, Y, 1),
    not(fact([_, _, b, Y])), not(fact([_, _, c, Y])),
    not(fact([_, _, d, Y])), opposite(Color, Opp),
    not(controlled(Opp, _, _, _, c, Y)),
    not(controlled(Opp, _, _, _, d, Y)).
owned(C, C):- player(C).
changeturn(Color):- 
    opposite(Color, Opp), delete([Color]), create([Opp]).
nocastle(Color, Man, X, Y):- 
    opposite(Color, Opp), fact([Opp, king, XK, YK]), 
    controlled(Color, Man, X, Y, XK, YK), delete([Opp, fortifiable]).
nocastle(Color, _, X, Y):- opposite(Color, Opp),
    fact([Opp, king, X, Y]), delete([Color, fortifiable]).
nocastle(Color, king, _, _):- delete([Color, fortifiable]).
nocastle(_, _, _, _).
do([Color, Man, X, Y, NX, NY]):- fact([C, M, NX, NY]),
    changeturn(Color), nocastle(Color, Man, NX, NY),
    delete([C, M, NX, NY]), delete([Color, Man, X, Y]), 
    create([Color, Man, NX, NY]).
do([Color, Man, X, Y, NX, NY]):- changeturn(Color),
    delete([Color, Man, X, Y]), nocastle(Color, Man, NX, NY),
    create([Color, Man, NX, NY]).
do([Color, Man, X, Y, NX, NY, New]):- changeturn(Color),
    delete([Color, Man, X, Y]), nocastle(Color, Man, NX, NY),
    create([Color, New, NX, NY]).
do([Color, castle, right, Y]):- changeturn(Color),
    delete([Color, king, e, Y]), delete([Color, rook, h, Y]), 
    delete([Color, fortifiable]),
    create([Color, king, g, Y]), create([Color, rook, f, Y]).
do([Color, castle, left, Y]):- changeturn(Color),
    delete([Color, king, e, Y]), delete([Color, rook, a, Y]), 
    delete([Color, fortifiable]),
    create([Color, king, d, Y]), create([Color, rook, c, Y]).
payoff([Color], 1.0):- 
    opposite(Color, Opp), todelete([Opp, king, _, _]).
payoff([Color], -1.0):- todelete([Color, king, _, _]).
\end{verbatim}
\bibliographystyle{splncs}
\bibliography{sidl3}

\begin{thebibliography}{10}

\bibitem{yepkio2019}
Yepkio:
\newblock Strategic interaction description language.
\newblock \url{https://github.com/Yepkio/sidl} (2019)

\bibitem{friedrich}
Michael, F., Ignatov, D.I.:
\newblock General game playing b-to-b price negotiations.
\newblock In: Proceedings of the Fifth Workshop on Experimental Economics and
  Machine Learning ({EEML 2019}) co-located with the 7th International
  Conference on Applied Research in Economics ({iCare7 2019}). Volume 2479 of
  {CEUR} Workshop Proceedings., Perm, Russia, CEUR-WS.org (2019)

\bibitem{gala}
Koller, D., Pfeffer, A.:
\newblock Representations and solutions for game-theoretic problems.
\newblock Artificial Intelligence \textbf{94} (1997)  167--215

\bibitem{ggp}
Genesereth, M.R., Love, N., Pell, B.:
\newblock General game playing: Overview of the aaai competition.
\newblock AI Magazine \textbf{26}(2) (2005)  62--72

\bibitem{tagiew2011beyond}
Tagiew, R.:
\newblock Beyond analytical modeling, gathering data to predict real agents'
  strategic interaction.
\newblock SKAD'11 -- Soft Computing Applications and Knowledge Discovery (2011)
   113

\bibitem{Balietti2017}
Balietti, S.:
\newblock nodegame: Real-time, synchronous, online experiments in the browser.
\newblock Behavior Research Methods \textbf{49}(5) (Oct 2017)  1696--1715

\bibitem{tagiewggma}
Tagiew, R.:
\newblock General game management agent.
\newblock CoRR \textbf{abs/0903.0353} (2009)

\bibitem{tagiewPhD}
Tagiew, R.:
\newblock {S}trategische {I}nteraktion realer {A}genten: {G}anzheitliche
  {K}onzeptualisierung und {S}oftwarekomponenten einer interdisziplin{\"a}ren
  {F}orschungsinfrastruktur.
\newblock PhD thesis, TU Bergakademie Freiberg (2011)

\bibitem{strips}
Fikes, R., Nilsson, N.:
\newblock Strips: a new approach to the application of theorem proving to
  problem solving.
\newblock Artificial Intelligence \textbf{2} (1971)  189--208

\bibitem{situationcalculus}
McCarthy, J., Hayes, P.J.:
\newblock Some philosophical problems from the standpoint of artificial
  intelligence.
\newblock Machine Intelligence \textbf{4} (1969)  463--502

\bibitem{pddl}
McDermott, D.V.:
\newblock The 1998 ai planning systems competition.
\newblock AI Magazine \textbf{21} (2000)  35--55

\bibitem{isoprolog}
Deransart, P., Ed-Dbali, A., Cervoni, L.:
\newblock Prolog: The Standard.
\newblock Springer-Verlag (1996)

\end{thebibliography}
\end{document}